\pdfoutput=1

\documentclass[11pt]{article}

\usepackage[preprint]{acl}

\usepackage{times}
\usepackage{latexsym}

\usepackage[T1]{fontenc}

\usepackage[utf8]{inputenc}

\usepackage{microtype}

\usepackage{inconsolata}

\usepackage{graphicx}

\usepackage{amsmath}
\definecolor{mygreen}{RGB}{34,139,34}
\definecolor{myyellow}{RGB}{204,153,0}
\usepackage{multirow}
\usepackage{makecell}
\usepackage{tcolorbox}

\title{%
  Advancing Question Generation with Joint Narrative and Difficulty Control\\
  \vspace{0.5em}
  {\large\textit{Preprint. Accepted to the BEA 2025 Workshop (ACL)}}
}

\author{First Author \\
  Affiliation / Address line 1 \\
  Affiliation / Address line 2 \\
  Affiliation / Address line 3 \\
  \texttt{email@domain} \\\And
  Second Author \\
  Affiliation / Address line 1 \\
  Affiliation / Address line 2 \\
  Affiliation / Address line 3 \\
  \texttt{email@domain} \\}

\author{
  Bernardo Leite \and Henrique Lopes Cardoso \\
  LIACC, Faculdade de Engenharia, Universidade do Porto \\
  Rua Dr. Roberto Frias, 4200-465 Porto, Portugal \\
  \texttt{\{bernardo.leite, hlc\}@fe.up.pt}
}

\begin{document}
\maketitle
\begin{abstract}
Question Generation (QG), the task of automatically generating questions from a source input, has seen significant progress in recent years. Difficulty-controllable QG (DCQG) enables control over the difficulty level of generated questions while considering the learner's ability. Additionally, narrative-controllable QG (NCQG) allows control over the narrative aspects embedded in the questions. However, research in QG lacks a focus on combining these two types of control, which is important for generating questions tailored to educational purposes. To address this gap, we propose a strategy for Joint Narrative and Difficulty Control, enabling \textit{simultaneous} control over these two attributes in the generation of reading comprehension questions. Our evaluation provides preliminary evidence that this approach is feasible, though it is not effective across all instances. Our findings highlight the conditions under which the strategy performs well and discuss the trade-offs associated with its application.
\end{abstract}

\section{Introduction}

\begin{figure}[ht]
\centering
\footnotesize
\begin{tcolorbox}[
    width=\columnwidth,
    colframe=black,       
    colback=gray!10,      
    arc=5mm,              
    boxrule=0.8pt,        
    left=6pt, right=6pt,  
    top=6pt, bottom=6pt,  
]
        \begin{flushleft}
        \textbf{Passage}: Once there were a hare and a turtle. The hare was proud of his speed and challenged the turtle to a race. Although the turtle was slow, he accepted. The hare quickly left the turtle behind but decided to rest and fell asleep. Meanwhile, the turtle kept going steadily and eventually reached the finish line first, winning the race.
        
        \hrulefill 

        \textbf{Narrative}: ``character'' \ \ \ \ \textbf{Difficulty}: ``\textcolor{mygreen}{easy}'' \newline
        \textbf{Generated QA Pair}: Who challenged the turtle to a race?' The hare

        \hrulefill 
        
        \textbf{Narrative}: ``outcome'' \ \ \ \ \textbf{Difficulty}: ``\textcolor{myyellow}{medium}'' \newline
        \textbf{Generated QA Pair}: What happened after the hare left the turtle behind? Decided to rest and fell asleep.

        \hrulefill 
        
        \textbf{Narrative}: ``outcome'' \ \ \ \ \textbf{Difficulty}: ``\textcolor{red}{hard}'' \newline
        \textbf{Generated QA Pair}: What happened because the turtle kept going steadily? The turtle won the race.

        \end{flushleft}
\end{tcolorbox}
\caption{Illustrative example of controlled question-answer generation with varying difficulty levels and narrative attributes.}
\label{fig:main_example}
\end{figure}

Question Generation (QG) focuses on the automated generation of coherent and meaningful questions targeting a data source, including unstructured text or knowledge bases \cite{rus2008question}.
Controllable QG plays a crucial role in education \cite{kurdi_2020_education}, as it facilitates the generation of personalized questions that address the unique needs and learning goals of students.
Recent work on QG utilized techniques such as fine-tuning \cite{zhang_2021_qg_survey,ushio_2022_qg_t5} and few-shot prompting \cite{wang_2022_towards_aied,chen_2024_storysparkqa} to generate questions based on a source text and, optionally, a target answer.
In controllable QG, this process is augmented by incorporating controllability labels into the input or prompt to guide the generation process. Specifically, research on Narrative-Controlled Question Generation (NCQG) focuses on controlling the \textbf{content} of generated questions, guided by underlying narrative elements (e.g., causal relationship) \cite{zhao_2022_cqg_acl,leite_2023_cqg_aied,zhang_2024_plans}. In turn, Difficulty-Controllable Question Generation (DCQG) emphasizes controlling the expected difficulty in answering the questions \cite{yifangao_2019_dqg,kumar_2019_dcqg,cheng_2021_dcqg,bi_2021_dcqg}. Some studies have considered the relationship between question \textbf{difficulty} and the \textbf{learner's ability} \cite{uto_2023_dif,tomikawa_2024_difmcqg}.

However, research in controllable QG lacks the combination of these two types of control, which is especially important to facilitate human control \cite{wang_2022_qgrecommendations} in the ever-increasing usage of generative models in this field.
Therefore, this research proposes a strategy that explores the feasibility of joining narrative and difficulty control to generate reading comprehension question-answer (QA) pairs from children-targeted narrative stories. Figure~\ref{fig:main_example} shows an example of the strategy.
Formally, we investigate the following research question (RQ): \textit{How effectively can we control the generation of question-answer pairs conditioned on both narrative and difficulty attributes using a modest\footnote{<1 billion of parameters.} scale model?}

For our experiments, we use a well-known dataset --- FairyTaleQA \cite{xu_2022_fairytaleqa} --- in which each question is already annotated with one of seven narrative labels.
Our method involves two main steps: (1) using simulated-learner QA systems to answer questions from FairyTaleQA, thereby estimating the difficulty labels via Item Response Theory, and (2) applying a joint narrative and difficulty control model, utilizing human-annotated narrative labels and the estimated difficulty labels for each question.

The proposed method is evaluated to determine whether both NCQG and DCQG have been successfully applied to the generated questions. For NCQG, we compare the similarity between human-authored and generated questions. For DCQG, we assess the performance of simulated-learner QA systems on questions generated with distinct difficulty levels. 
Although the results demonstrate the effectiveness of the strategy, NCQG shows consistent success, whereas DCQG exhibits moderate success, with performance varying across specific narrative attributes and difficulty levels. Our goal is to highlight the conditions under which the strategy performs with high or low efficacy, providing insights for researchers pursuing similar research lines.
In summary, our contributions are:
\begin{itemize}
\item We propose a joint strategy for controlling the generation of question-answer pairs conditioned on narrative and difficulty attributes.
\item We report on the linguistic features influenced by control and conduct an error analysis of the generated QA pairs, providing insights into the performance and limitations of the method.
\end{itemize}

\section{Background and Related Work}

\subsection{Controllable Question Generation (CQG)} \label{sec:sota_cqg}

As stated by \citet{zhang_2024_plans}, prior research on CQG has explored two main perspectives: content (or type) and difficulty.

\textbf{Content control} relates to the linguistic elements incorporated into the generated questions. For instance, \citet{ghanem_2022_cqg_acl} proposed controlling specific reading comprehension skills, such as figurative language and vocabulary. Additionally, \citet{zhao_2022_cqg_acl} focused on controlling narrative elements, while \citet{leite_2023_cqg_aied} extended this approach by controlling explicitness attributes. \citet{elkins_2023_cqg_aied} propose to control Bloom's question taxonomy \cite{bloom_2002_revised}.

\textbf{Difficulty control} is related to the challenge of answering the generated questions, a concept that is often subjective (i.e., difficulty can vary depending on the respondent). In this regard, \citet{yifangao_2019_dqg} assigned difficulty labels (easy or hard) to questions based on whether QA systems could answer them correctly and used these labels as inputs to control the generation process. \citet{kumar_2019_dcqg} proposed estimating difficulty based on named entity popularity, while \citet{bi_2021_dcqg} tackle the challenge of high diversity in QG. Furthermore, \citet{cheng_2021_dcqg} controlled question difficulty by considering the number of inference steps required to arrive at an answer.

One limitation of previous approaches is (1) the lack of emphasis on the relationship between question difficulty and learner ability. Addressing this problem, \citet{uto_2023_dif} proposed to use Item Response Theory (IRT) \cite{lord_2012_irt}, a mathematical framework in test theory, to quantify question difficulty and directly relate it to learner ability. Another limitation is (2) the lack of integration of multiple attributes. While \citet{zhang_2024_plans} combine both narrative and difficulty attributes, they define \textit{difficulty} in terms of answer explicitness and the number of sentences needed to answer the questions.
The novelty of this study lies in integrating content control, through narrative elements, with difficulty control \textit{informed by simulated learners' ability}, thus building on the foundations laid by previous research.

\subsection{Item Response Theory (IRT)} \label{sec:sota_irt}

IRT \cite{lord_2012_irt} is a statistical framework used to study the interaction between test-takers (ability or proficiency) and their performance on test items. A key aspect of IRT is to model the relationship between question difficulty and learner ability, offering insights into how well a question differentiates between individuals with varying levels of skill. This relationship allows for an estimation of the likelihood that a learner with a specific ability level can correctly answer a given question, making it particularly useful for adaptive testing and understanding question complexity.
A commonly used model in IRT is the \textbf{Rasch model}, which assumes that the probability of a correct response depends on the relation between learner ability ($\theta$) and the item's difficulty ($b$):

\begin{equation}
P(X_{ij} = 1 \mid \theta_i, b_j) = \frac{e^{\theta_i - b_j}}{1 + e^{\theta_i - b_j}},
\end{equation}
where $\theta_i$ is the learner ability of individual $i$, $b_j$ is the difficulty of item $j$, and $P(X_{ij} = 1 \mid \theta_i, b_j)$ is the probability that individual $i$ correctly answers item $j$. In our study, we use IRT to estimate both question difficulty (\(b\)) and learner ability (\(\theta\)) parameters.

\subsection{FairyTaleQA: Purpose and Value} \label{sec:sota_fairytaleqa}
We use the FairytaleQA dataset \cite{xu_2022_fairytaleqa} because its stories and corresponding question-answer pairs align with the goal of addressing \textit{narrative comprehension}. According to \citet{xu_2022_fairytaleqa}, narrative comprehension represents a high-level cognitive skill closely linked to overall reading proficiency \cite{lynch_2008_reading}.
A key feature of FairytaleQA is the expert annotations on each question, which are grounded in evidence-based frameworks \cite{paris_2003_narrative,alonzo_2009_narrative}. The annotated narrative elements targeted for control are:

\begin{itemize}
\item \textbf{Setting}: Focusing on the time and place of events, often starting with ``Where...?'' or ``When...'';
\item \textbf{Action}: Related to the actions of characters;
\item \textbf{Feeling}: Exploring emotional states or reactions (e.g., ``How did/does X feel?'');
\item \textbf{Causal relationship}: Addressing cause-and-effect (e.g., ``Why...?'' or ``What caused/made X?'');
\item \textbf{Outcome resolution}: Focusing on the outcomes of events (e.g., ``What happened/happens after X?'');
\item \textbf{Prediction}: Questions about future or unknown events based on textual evidence.
\end{itemize}

While there are other popular educational QA datasets (following the open-ended \textit{wh}-questions format), such as NarrativeQA \citep{kocisky_2018_narrativeqa} and StoryQA \citep{zhao_storyqa_2023}, they are not annotated with specific reading comprehension skills. This further motivated our decision to use FairytaleQA in this study.

\section{Method}

\begin{figure*}[!ht]
    \centering
    \includegraphics[width=\textwidth]{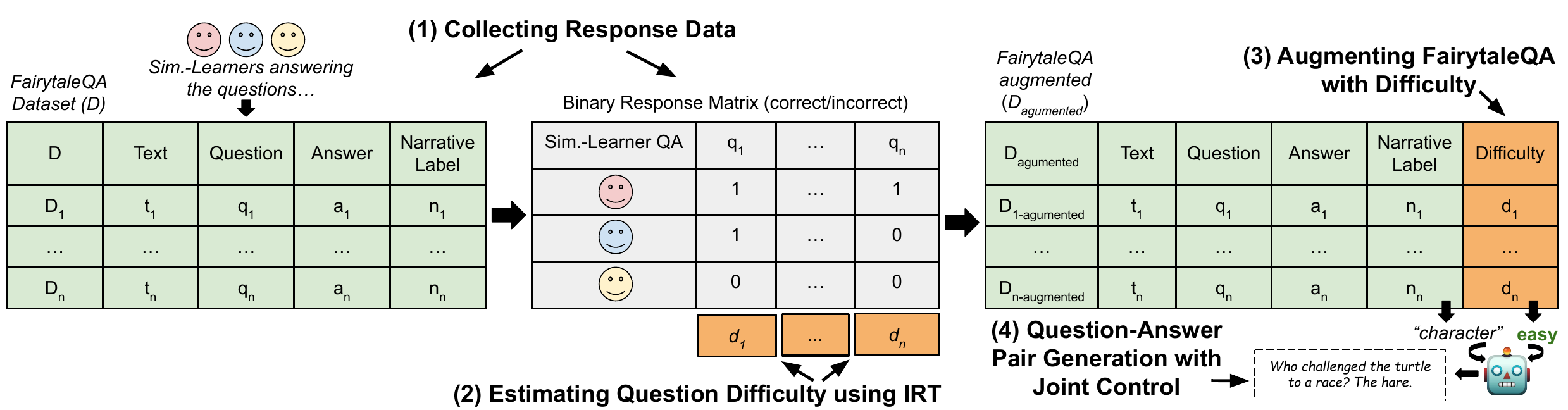}
    \caption{Overall methodology for joint narrative and difficulty control.}
    \label{fig:overall_method}
\end{figure*}

This section outlines the methodology of this research, which includes augmenting FairytaleQA with IRT-based difficulty labels and developing a question-answer pair generation model with joint narrative and difficulty control.
Figure~\ref{fig:overall_method} provides an overview of the steps discussed in this section.

\subsection{Augmenting FairytaleQA With IRT-Based Question Difficulty Labels} \label{sec:augment_fairytale}

Let $D$ be our dataset consisting of instances represented as quartets:
\begin{equation}
D_{i} = (t, q, a, n),
\end{equation}
where $t$ is a text, $q$ is the question, $a$ is the answer about the text, and $n$ is the narrative element associated with the question-answer pair $(q, a)$. The aim is to create a fifth element $d$, resulting in a new instance augmented:
\begin{equation}
D_{i-\text{augmented}} = (t, q, a, n, d),
\end{equation}
where $d$ is the estimated difficulty value associated with the question-answer pair $(q, a)$. To create these augmented instances, we used the method proposed by \citet{uto_2023_dif} and \citet{tomikawa_2024_adpative}:
\begin{enumerate}
    \item \textbf{Collecting response data for each question-answer pair}: We collected answers to the questions from multiple respondents. Given the unavailability of real students, we utilized simulated-learner QA systems, which are models capable of automatically extracting answers to the posed questions. As explained in Section~\ref{sec:creating_qa_systems}, the QA models were deliberately chosen to represent different levels of performance to simulate varying ability levels.
    \item \textbf{Estimating Question Difficulty with IRT}: Using the answers collected from the simulated-learner QA systems, we estimated the difficulty of each question using IRT, specifically employing the Rasch model as described in Section~\ref{sec:sota_irt}.
    \item \textbf{Augmenting FairytaleQA with difficulty estimates}: Based on the estimated difficulty values, we augment each instance of the dataset with $d$, resulting in $D_{i-\text{augmented}} = (t, q, a, n, d)$.
\end{enumerate}

\subsection{Question-Answer Pair Generation with Joint Narrative and Difficulty Control}

The controllable process can be represented as follows: given an instruction prompt $p$, the aim is to use a model $M$ to generate a question-answer pair $(q_{\text{new}}, a_{\text{new}})$. This can be formulated as:
\begin{equation}
(q_{\text{new}}, a_{\text{new}}) = \text{M}(p),
\end{equation}
where prompt $p$ incorporates the desired narrative label $n$, difficulty value $d$, and target text $t$. The prompt follows this template:
\begin{quote}
``Generate a $\langle{d}\rangle$ question-answer pair about narrative label $\langle{n}\rangle$ considering the following text: $\langle{t}\rangle$''
\end{quote}
$\text{M}$ is an encoder-decoder model that is fine-tuned using \allowbreak
$D_{i-\text{augmented}} = (t, q, a, n, d)$ instances. The encoder receives prompt $p$ and encodes it into a fixed-length representation known as a context vector. The decoder takes the context vector and generates the output text $(q_{\text{new}}, a_{\text{new}})$, using special tokens \textsc{$\langle$QU$\rangle$} and \textsc{$\langle$AN$\rangle$} that serve to differentiate between $q_{\text{new}}$ and $ a_{\text{new}}$. The idea is to guide the model in generating a question-answer pair of the intended difficulty $d$ and narrative element $n$.

\section{Experimental Setup}

\subsection{Preparing the FairytaleQA Dataset} \label{sec:preparing_fairy_data_setups}

We use FairyTaleQA \cite{xu_2022_fairytaleqa}, which comprises 10,580 question–answer pairs manually created by educational experts based on 278 narrative stories. Each story contains approximately 15 section texts, and each section (about 149 tokens) contains approximately 3 question–answer pairs.
From the original dataset, we have prepared different data setups\footnote{The arrow separates the input (left) and output (right) information. On the left part, the + symbol illustrates whether the method incorporates control attributes.} for generating a QA pair:
\begin{itemize}
  \item \textbf{Text $\rightarrow$ QA}: This setup only contains the text as input, so it serves as a baseline to compare with the subsequent setups, which consider control attributes.
  \item \textbf{Nar + Text $\rightarrow$ QA}: This setup considers \textit{narrative} as a control attribute in the input.
  \item \textbf{Dif + Text $\rightarrow$ QA}: This setup considers \textit{difficulty} as a control attribute in the input.
  \item \textbf{Nar + Dif + Text $\rightarrow$ QA}: This setup considers both the narrative and difficulty attributes.
\end{itemize}

\subsection{Creating Simulated-Learner QA Systems} \label{sec:creating_qa_systems}

To create the simulated-learner QA systems, we trained five QA models.
The choice of five was made empirically: it provided sufficient granularity for analysis while avoiding ties that could arise with fewer levels (e.g., four).
The selected encoder models are \texttt{DeBERTaV3} \cite{he_2021_debertav3}, \texttt{RoBERTa} \cite{liu_2019_roberta}, \texttt{BERT} \cite{devlin_2019_bert} and \texttt{DistilBERT} \cite{sanh_2019_distilbert}. We also use one decoder: \texttt{GPT-2} \cite{radford_2019_gpt2}. They were fine-tuned on separate general-purpose question answering data (the SQuAD v1.1 dataset \cite{rajpurkar_2016_squad}). The models were deliberately chosen for their varying performance levels, thereby simulating different levels of learner skill. 
Table~\ref{tab:qa_students_performance} shows the performance of each QA system on the SQuAD v1.1 evaluation set, using the \textit{n}-gram similarity metric ROUGE$_\text{L}$-F1 \cite{lin_rouge_2004} (QA answer vs. SQuAD ground-truth answer).
\begin{table}[!ht]
\centering
\renewcommand{\arraystretch}{1.2} 
\caption{Simulated-Learner QA systems performance on SQuAD v1.1 evaluation set.}
\label{tab:qa_students_performance}
\begin{tabular}{lc} \hline
\textbf{Sim.-Learner QA} & \textbf{ROUGE$_L$-F1 (0-1)} \\ \hline
\texttt{DeBERTaV3} (large) & 0.87 \\
\texttt{RoBERTa} (base) & 0.82 \\
\texttt{BERT} (base) & 0.75 \\
\texttt{DistilBERT} (base) & 0.69 \\
\texttt{GPT-2} & 0.46 \\ \hline
\end{tabular}
\end{table}

\subsection{Answering FairytaleQA Questions with QA Systems}

For each question in the train and validation sets of the FairyTaleQA dataset, all five simulated-learner QA systems generated their own answers. Each QA answer is then compared to the corresponding ground-truth answer to determine correctness. We considered an answer correct if it achieved either an exact match score of 1 or a ROUGE$_L$-F1 score of at least 0.5.
The QA answers are organized into a binary response matrix --- Figure~\ref{fig:overall_method} shows an example of such a matrix. Each row corresponds to a simulated-learner QA system and each column corresponds to a question ID. Each cell contains a 0 or 1, indicating incorrect or correct answers, respectively. This matrix serves as input data for the subsequent question difficulty estimation using IRT.

\subsection{Estimating Question Difficulty with IRT}

Based on the collected correct and incorrect answers for each question --- organized into a binary response matrix --- we estimated question difficulty using the Rasch Model (recall Section~\ref{sec:sota_irt}). Specifically, using the binary correctness data produced by the simulated-learner QA systems, the estimation is performed using the Expectation-Maximization (EM) algorithm \cite{embretson_irt_2000}. This yielded difficulty values that were subsequently normalized to a 0-1 scale (0, 0.28, 0.50, 0.72, and 1), where higher values represent more difficult questions. The numerical values were converted into corresponding categorical labels -- \textit{easy}, \textit{medium}, \textit{moderate}, \textit{hard}, and \textit{extreme} -- to be used in textual prompts. The distribution of the estimated difficulty values by narrative label in the data is presented in Table~\ref{tab:dist_values_train_val}. Some attributes (e.g., \textit{feeling} and \textit{prediction}) have limited representation in the dataset.

\begin{table}
\centering
\renewcommand{\arraystretch}{1.2} 
\begin{tabular}{c|ccccc} \hline
\textbf{Nar.} & Easy & Med. & Mod. & Hard & Extr. \\ \hline
Action & 773 & 362 & 375 & 435 & 749 \\
Causal & 316 & 200 & 245 & 316 & 1291 \\
Char. & 497 & 133 & 101 & 116 & 115 \\
Feeling & 55 & 79 & 62 & 89 & 539 \\
Out. & 126 & 114 & 138 & 165 & 268 \\
Pred. & 22 & 21 & 23 & 50 & 250 \\
Setting & 276 & 70 & 60 & 54 & 63 \\ \hline
Action & 76 & 40 & 65 & 60 & 92 \\
Causal & 35 & 27 & 31 & 50 & 151 \\
Char. & 50 & 17 & 14 & 9 & 17 \\
Feeling & 0 & 9 & 9 & 5 & 71 \\
Out. & 11 & 13 & 19 & 15 & 39 \\
Pred. & 1 & 3 & 6 & 7 & 38 \\
Setting & 29 & 4 & 5 & 4 & 3 \\ \hline
\end{tabular}
\caption{Difficulty values by Nar. (train and val set).}
\label{tab:dist_values_train_val}
\end{table}

Additionally, using the Maximum a Posteriori (MAP) algorithm \cite{embretson_irt_2000}, we estimated the ability ($\theta$) values for each QA system. These values are reported in Table~\ref{tab:qa_ability_values}, with higher values representing higher abilities. These values align, as expected, with the systems' original performance levels shown in Table~\ref{tab:qa_students_performance}.

\begin{table}[!ht]
\centering
\renewcommand{\arraystretch}{1.2} 
\begin{tabular}{lc} \hline
\textbf{Sim.-Learner QA} & \textbf{Ability ($\theta$)} \\ \hline
\texttt{DeBERTaV3} (large) & 0.43 \\
\texttt{RoBERTa} (base) & 0 \\
\texttt{BERT} (base) & -0.66 \\
\texttt{DistilBERT} (base) & -1.25 \\
\texttt{GPT-2} & -1.60 \\ \hline
\end{tabular}
\caption{Simulated-learner estimated ability values ($\theta$) after answering questions from the FairytaleQA dataset.}
\label{tab:qa_ability_values}
\end{table}

We use \textit{mirt}\footnote{\url{https://cran.r-project.org/web/packages/mirt/index.html}} tool for IRT, including all estimations.

\subsection{Creating a Question-Answer Pair Generation Model} \label{sec:creating_qg_model}

We use the \texttt{Flan-T5} \cite{chung_2024_flant5} encoder-decoder model for the controllable task. This model builds upon the original \texttt{T5} \cite{raffel_2020_t5}, which has been fine-tuned with task-specific instructions using prefixes, making it well-suited for our methodology. Additionally, \texttt{Flan-T5} demonstrates remarkable performance in text generation tasks, particularly in QG \cite{chen_2024_storysparkqa, zhang_2024_plans}. We employ the \texttt{flan-t5-large} version, which is publicly available via Hugging Face\footnote{\url{https://huggingface.co/google/flan-t5-large}}.
Training is conducted for up to 10 epochs, with early stopping implemented using a patience of 2 epochs.
During inference, we apply Top-k sampling with $k=50$, $p=0.9$ and $temp=1.2$ to encourage diversity (values obtained experimentally). We initially explored beam search, a widely used technique in QG; however, we observed that it frequently produced repetitive questions when tasked with generating questions for the same narrative element across different difficulty levels.

\subsection{Generating QA Pairs for Evaluation} \label{sec:gen_qas_test}

We fine-tune the \texttt{Flan-T5} model on the training set of FairytaleQA. We obtain 4 models, as the model has been trained on each of the 4 data setups described in Section~\ref{sec:preparing_fairy_data_setups}. For the 2 setups where difficulty labels are used, we apply the resulting models (inference) to the corresponding test set and generate 5 QA pairs for each text's section --- one QA pair for each difficulty label. Since the FairytaleQA test set contains 394 section texts, we obtain a total of 1,970 generated QA pairs. Additionally, each text includes human-authored QA pairs associated with different narrative labels. This approach ensures that the generated QA pairs are balanced across distinct difficulty levels and narrative elements for further evaluation.

\section{Evaluation}

\subsection{Evaluation Procedure}

For NCQG, our evaluation protocol follows prior studies \cite{zhao_2022_cqg_acl,leite_2023_cqg_aied,leite_2024_fcqg_csedu} that focused on controlled generation using narrative labels. For DCQG, the evaluation protocol is based on recent works \cite{uto_2023_dif,tomikawa_2024_adpative,tomikawa_2024_difmcqg} that emphasize the use of simulated-learner QA systems across generated questions with distinct difficulty levels.

\textbf{Narrative Control}: To assess narrative control, we use a standard approach in QG: comparing generated questions directly with human-authored ground-truth questions. Hypothesis 1 (H1) is that \textit{incorporating narrative attributes will result in generated questions that are more similar to the ground-truth}, as previously shown by \citet{leite_2024_fcqg_csedu}.
To quantify the similarity, we employ the \textit{n}-gram similarity metric ROUGE$_L$-F1 \cite{lin_rouge_2004}, as originally adopted by the FairytaleQA authors.
For a better perception of the idea, consider the human-authored ground-truth question: ``What did Matte and Maie do on Saturdays?'' (annotated with the \textit{setting} narrative element) and the generated question targeting the same narrative element: ``What did Maie and Matte do to provide for themselves?''. These questions yield a high ROUGE$_L$-F1 score because they are similar in terms of the narrative-related vocabulary they share, thus indicating successful narrative control.

\textbf{Difficulty Control}: For difficulty control, the evaluation focuses on analyzing the performance of simulated-learner QA systems when answering questions generated at varying difficulty levels. Hypothesis 2 (H2) posits that \textit{simulated-learner QA systems will perform better on easier questions and worse on more difficult ones, relative to their ability levels}.

\subsection{Results} \label{sec:results}

\begin{table*}[!ht]
\centering
\renewcommand{\arraystretch}{1.2} 
\begin{tabular}{l cccccc c} \hline
\textbf{Data Setup} & \textbf{Char.} & \textbf{Setting} & \textbf{Action} & \textbf{Feeling} & \textbf{Causal} & \textbf{Out.} & \textbf{Pred.} \\ \hline
\textbf{Text $\rightarrow$ QA} & .227 & .269 & .287 & .281 & .271 & .227 & .251 \\
\textbf{Nar + Text $\rightarrow$ QA} & .304 & .537 & .427 & .527 & .412 & .458 & .348 \\
\textbf{Nar + Dif + Text $\rightarrow$ QA} & .305 & .530 & .412 & .529 & .405 & .425 & .365 \\ \hline
\end{tabular}
\caption{\textbf{Narrative Control}: Similarity (ROUGE$_L$-F1) between generated and ground-truth questions on the test set by narrative element.
\textbf{Text $\rightarrow$ QA} is used as a baseline to assess whether narrative control helps the generated questions approximate the ground-truth questions.}
\label{tab:results_narrative_control}
\end{table*}

\textbf{Narrative Control}: Table~\ref{tab:results_narrative_control} presents the results from the narrative control perspective, measured using ROUGE$_L$-F1 \textit{n}-gram similarity between the generated questions and the human-authored ground-truth questions.
We observe an improvement in the similarity to ground-truth questions when narrative control attributes are incorporated. This trend is consistently observed across all seven narrative labels. Furthermore, these findings align with the results reported in prior studies on narrative control \cite{leite_2023_cqg_aied,leite_2024_fcqg_csedu}.
Of novelty, when narrative and difficulty labels are fused, we observe a similar improvement trend, comparable to the incorporation of narrative attributes alone.
These results support Hypothesis 1 (H1), indicating that our method effectively controls the narrative elements underlying the generated questions. 
Appendix~\ref{sec:sem_control} shows further support by reporting semantic proximity results.

\begin{figure}[!ht]
    \centering
    \includegraphics[width=\columnwidth]{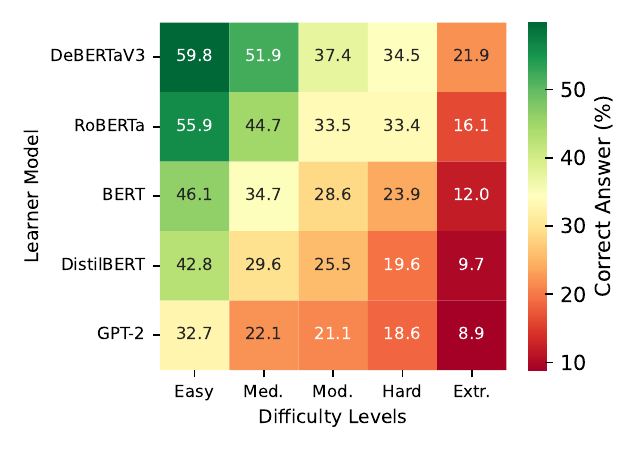}
    \caption{Percentage (\%) of correct answers by difficulty level when only difficulty control labels are used (\textbf{Dif + Text $\rightarrow$ QA}).}
    \label{fig:results_en_5dif}
\end{figure}

\textbf{Difficulty Control}: Figure~\ref{fig:results_en_5dif} presents the results for difficulty control only, showing the percentage of correct responses from the simulated-learner QA systems across all difficulty levels. The percentage of correct answers decreases as the difficulty level increases for all simulated learners\footnote{All percentages are relatively low ($<$60). This is because the QA models were not trained on the FairyTaleQA dataset but were instead trained on SQuAD. This intentional choice ensures that the models' knowledge remains unbiased with respect to FairyTaleQA content.}.
Additionally, learners with higher abilities achieve higher percentages of correct answers, while those with lower abilities achieve lower percentages. These findings are consistent with previous works \cite{uto_2023_dif,tomikawa_2024_adpative} and support Hypothesis 2 (H2), demonstrating that the method controls the difficulty levels of the generated questions.

\begin{figure}[!ht]
    \centering
    \includegraphics[width=\columnwidth]{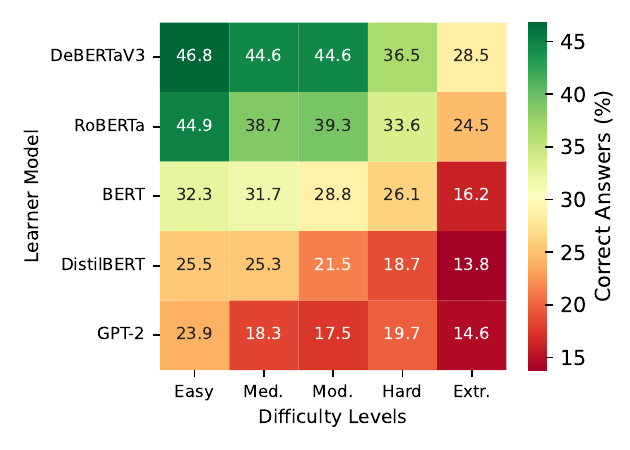}
    \caption{Percentage (\%) of correct answers by difficulty level when both difficulty and narrative control labels are used (\textbf{Nar + Dif + Text $\rightarrow$ QA}).}
    \label{fig:results_en_5dif_nar}
\end{figure}

\noindent \textbf{Joint Narrative and Difficulty Control}: Figure~\ref{fig:results_en_5dif_nar} presents the results for difficulty control when difficulty and narrative attributes are fused.
In most cases, the percentage of correct answers decreases as the difficulty level increases across all simulated learners. These findings demonstrate that even when conditioning the generation process on both narrative content and difficulty, it remains possible to perform difficulty control.
However, some inconsistencies are observed: for \texttt{DeBERTaV3}, there is no distinction between medium and moderate difficulty levels; for \texttt{RoBERTa}, the percentage of correct answers increases between medium and moderate levels; and for \texttt{GPT-2}, a similar trend occurs between moderate and hard levels.
For an overall graphical comparison of difficulty control using only difficulty versus combining difficulty and narrative attributes, see Appendix~\ref{sec:results_en_dif_vs_difnar}.

Figure~\ref{fig:results_en_nar_by_5dif} shows the overall accuracy for each narrative label, with trends suggesting difficulty control particularly between easy, hard, and extreme levels. However, control becomes inconsistent at intermediate levels. Among the attributes, \textit{causal} and \textit{outcome} demonstrate the most consistent control across difficulty levels, while \textit{prediction} and \textit{feeling} exhibit the least success. This inconsistency can be related to the limited representation of these attributes in the FairytaleQA dataset (recall Table~\ref{tab:dist_values_train_val}), which prevents the model from learning to generate questions across different difficulty levels. Additionally, questions tied to these attributes are inherently more challenging, as reflected in the lower global performance of simulated-learner QA systems.
\begin{figure}[!ht]
    \centering
    \includegraphics[width=\columnwidth]{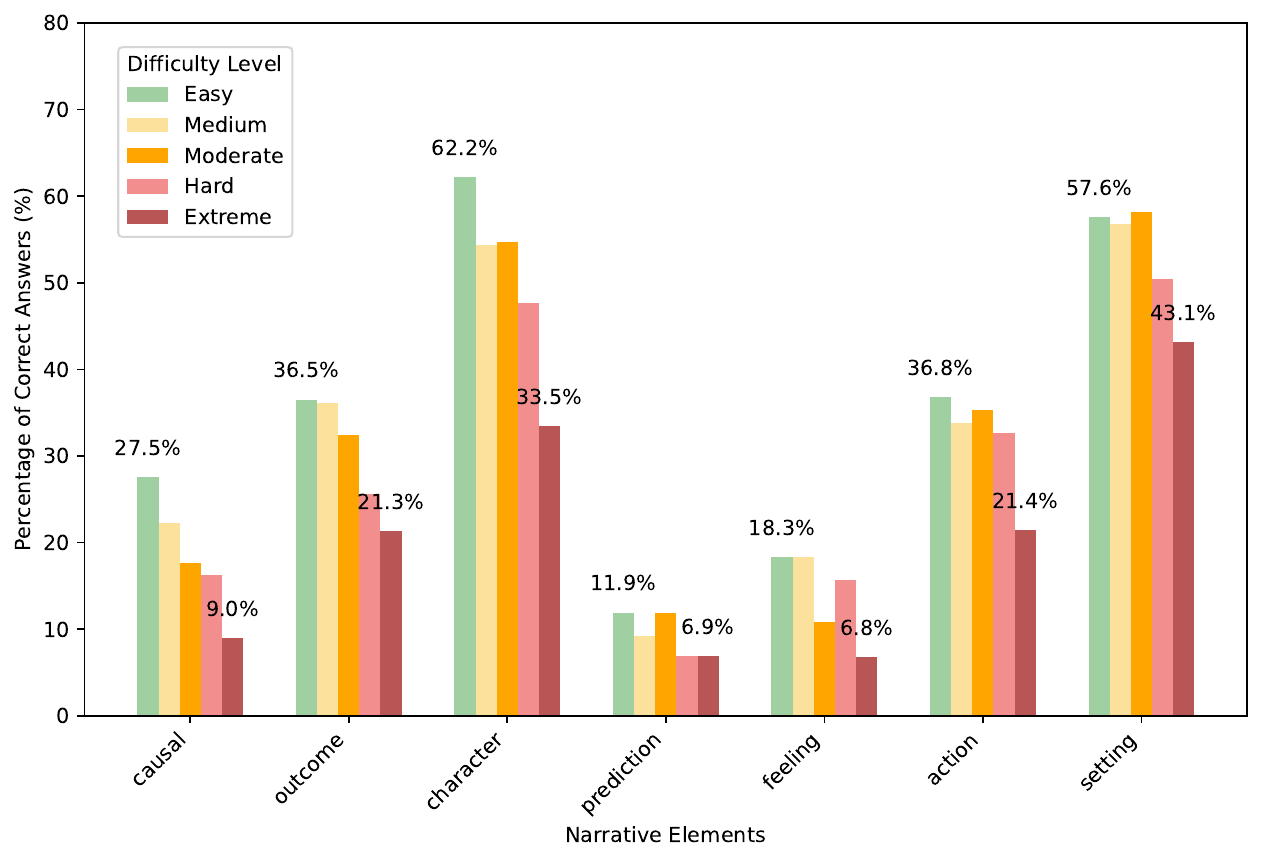}
    \caption{Percentage (\%) of correct answers per narrative element and difficulty level (5 levels).}
    \label{fig:results_en_nar_by_5dif}
\end{figure}
\begin{figure}[!ht]
    \centering
    \includegraphics[width=\columnwidth]{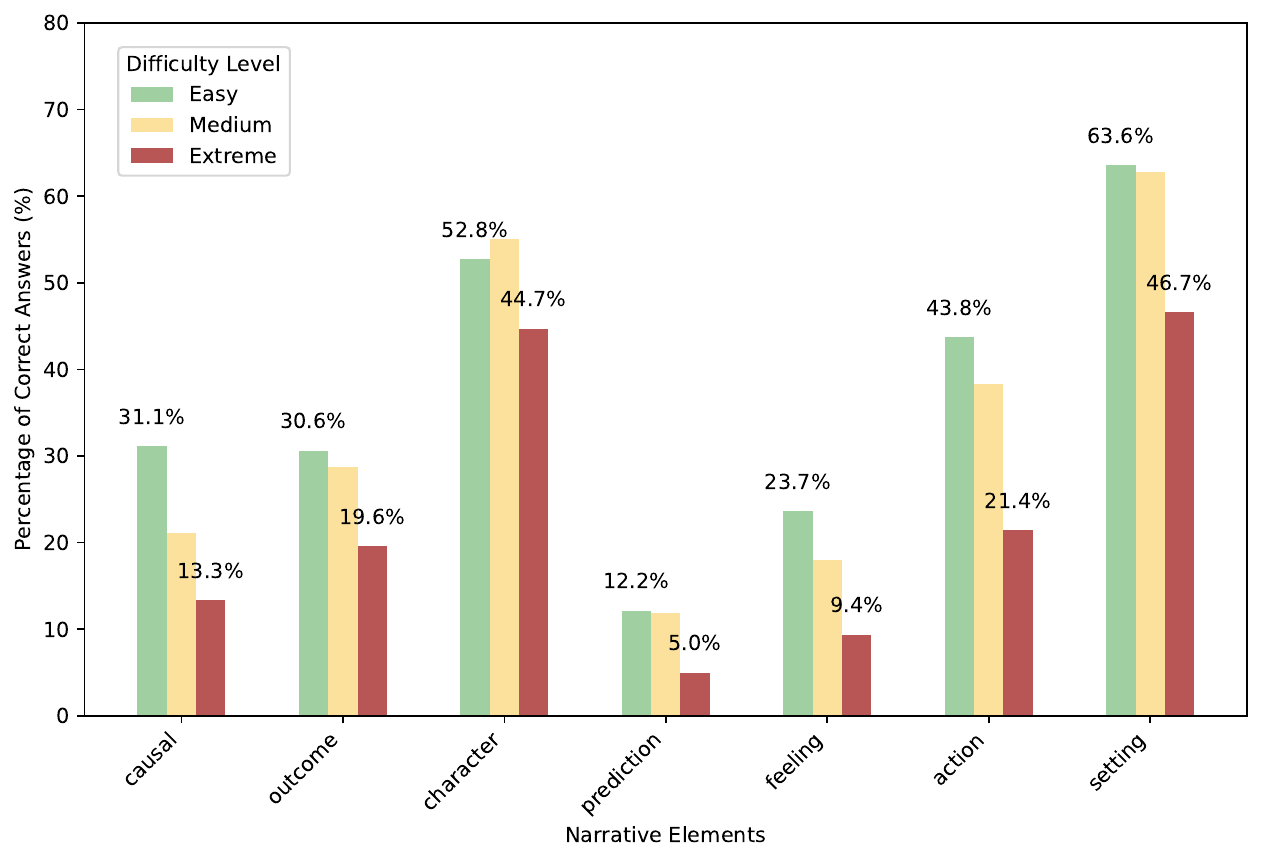}
    \caption{Percentage (\%) of correct answers per narrative element and difficulty level (3 levels).}
    \label{fig:results_en_nar_by_3dif}
\end{figure}
For attributes such as \textit{character}, \textit{prediction}, \textit{action}, and \textit{setting}, the confusion is particularly evident between medium and moderate levels. To address this, we experimented with an alternative model trained on a lower granularity of difficulty levels, combining medium, moderate, and hard into a single medium level. In Figure~\ref{fig:results_en_nar_by_3dif}, we show the result of this experiment, which demonstrates more consistent control across all levels. However, the \textit{character} and \textit{prediction} attributes continue to reveal some difficulty in distinguishing levels.
These results support Hypothesis 2 (H2), confirming that the joint method enables difficulty control, although with less consistency than when controlling for difficulty alone.
In Section~\ref{sec:discussion}, we outline potential explanations for these results.

\noindent \textbf{Linguistic Features Influenced By Control}:
To better understand the linguistic features influenced by the controllability method, we analyze the linguistic properties of the generated QA pairs across different difficulty levels and narrative attributes. Prior work on difficulty-only controlled generation \citep{tomikawa_2024_adpative} identifies two key factors that distinguish difficulty levels: (1) the average number of words in the generated answers, and (2) the distribution of initial interrogative terms in the generated questions.
While we also explore these features (see Appendix~\ref{sec:linguistic_features}), we emphasize here a novel aspect that we also found experimentally to be relevant: (3) the degree of lexical novelty in the generated QA pairs relative to the source narrative text.
To quantify this, we use the PINC (Paraphrase In N-gram Changes) metric \cite{chen_2011_pinc}, which computes the percentage of \textit{n}-grams present in the generated QA pairs but not in the source text.
Higher PINC scores indicate greater lexical novelty and diversity.
The results in Table~\ref{tab:pinc_scores} show that the diversity of the QA pairs increases with higher difficulty levels. This trend is observed both when difficulty labels are used independently and when combined with narrative labels. Therefore, we conclude that the linguistic diversity between the generated QA pairs and the source text is a feature influenced by difficulty control, regardless of whether difficulty labels are used alone or in conjunction with narrative labels.

\begin{table}[!ht]
\centering
\begin{tabular}{ccccc}
\hline
\textbf{Data Setup} & & \textbf{Easy} & \textbf{Med.} & \textbf{Extr.} \\ \hline
\multirow{2}{*}{\makecell[c]{\textbf{Dif + Text} \\ $\rightarrow$ \textbf{QA}}} & Q & 55.60 & 60.23 & 63.94 \\
\cline{2-5}
 & A & 9.88 & 23.17 & 48.69 \\ \hline
\multirow{2}{*}{\makecell[c]{\textbf{Nar + Dif + Text} \\ $\rightarrow$ \textbf{QA}}} & Q & 57.34 & 60.72 & 65.57 \\
\cline{2-5}
 & A & 22.02 & 26.00 & 41.14 \\ \hline
\end{tabular}
\caption{PINC values (\%) considering 3-gram novelty for generated questions (Q) and answers (A).}
\label{tab:pinc_scores}
\end{table}

\noindent\textbf{Error Analysis of Generated QA Pairs}:
We randomly selected 100 generated QA pairs and analyzed potential problems. Two main issues were identified, as shown in Figure~\ref{fig:error_analysis_examples}: (1) \textit{hallucinated content} (14\%) and (2) \textit{nonsensical QA pairs} (10\%).

In the first example (1), the generated QA pair introduces made-up terms that do not exist in the source text. This is the most frequent problem encountered, likely due to the nature of the inference method (Top-k sampling). 
In example 2a, the problem is that the generated QA pair misleading.
In Example 2b, the question seeks a motivation not in the text, and the vague answer just reuses the question's wording without adding meaning.

\begin{figure}[!ht]
\centering
\footnotesize
\noindent\fbox{%
    \parbox{\columnwidth}{
    {
        \begin{center}
        Example 1 (\textbf{hallucinated content}):
        \end{center}
        \textbf{Text}: And then he cried for the third time: ``Old Hopgiant!'' And there appeared a fellow of truly horrible size, with a scythe as large as a ship's mast.
        \newline
        \textbf{Gen. Q}: Who appeared to the \textcolor{red}{hunter} at the \textcolor{red}{swan}?
        \newline
        \textbf{Gen. A}: A fellow of truly horrible size, with a scythe as large as a ship's mast.
        \newline
        \textbf{\textit{Explanation}}: There is no mention of any hunter or swan in the source text.
        \begin{center}
        Example 2a (\textbf{nonsensical QA pairs}):
        \end{center}
        \textbf{Text}: When the old man had finished his dance, the big demon said: ``Thank you very much for your amusing dance. Now give us the pleasure of drinking a cup of wine with us,'' and with these words he handed him his largest wine-cup.
        \newline
        \textbf{Gen. Q}: What did the old man do after he finished dancing?
        \newline
        \textbf{Gen. A}: \textcolor{red}{Danced with the big demon}.
        \newline
        \textbf{\textit{Explanation}}: The old man had finished dancing and did not dance with anyone again.
        \begin{center}
        Example 2b (\textbf{nonsensical QA pairs}):
        \end{center}
        \textbf{Text}: But the rich neighbor wanted the meadow for himself alone, and told the poor one that he would drive him out of house and home if he did not come to an agreement with him that whichever one of them mowed the largest stretch of the meadowland in a single day, should receive the entire meadow.
        \newline
        \textbf{Gen. Q}: Why did the neighbor want to \textcolor{red}{own the meadow}?
        \newline
        \textbf{Gen. A}: He wanted to \textcolor{red}{have the meadow}.
        \newline
        \textbf{\textit{Explanation}}: The question contains the answer.
    }
    }%
}
\caption{Examples of problematic generated question-answer pairs (error analysis).}
\label{fig:error_analysis_examples}
\end{figure}

\section{Discussion} \label{sec:discussion}

We revisit our research question (RQ): \textit{How effectively can we control the generation of question-answer pairs conditioned on both narrative and difficulty attributes using a modest scale model?}

Our findings indicate that the model effectively controls narrative elements, whether alone or combined with difficulty attributes. It also shows effective difficulty control when conditioned solely on difficulty attributes. The joint model, which combines narrative and difficulty attributes, generally achieves consistent control for at least three levels (easy, hard, and extreme). However, inconsistencies arise in the intermediate levels (medium and moderate). We also observed that certain attributes are more conducive to effective control, while others, like \textit{prediction} and \textit{feeling}, are less effective. Notably, reducing the granularity of difficulty levels improves the overall control.
We now delve into two main factors that underlie our findings.

First, \textit{generating QA pairs while simultaneously controlling both difficulty and narrative attributes is an inherently challenging task}. When the narrative element is fixed, the space of plausible questions becomes more constrained. This makes it harder to vary difficulty meaningfully, as the questions tend to focus on similar content. For instance, in Figure~\ref{fig:main_example}, the last two questions share the same narrative element but differ in difficulty. This overlap in content makes it harder to generate questions with clearly distinct difficulty levels.

Second, \textit{some narrative attributes naturally lead to easier questions}. For instance, the \textit{character} attribute often involves straightforward ``Who'' questions, making it harder to create questions with distinct difficulty levels. In contrast, questions following the \textit{prediction} attribute are demanding, adding complexity to the learning process of generating well-differentiated questions.

\textbf{Transferability to other domains}: While our current work focuses on narrative comprehension, the principles of controllable QG are not domain-specific. For instance, it would be feasible to control generation based on other reading comprehension skills, as explored by \citet{ghanem_2022_cqg_acl}. Progress in this direction depends on the availability of datasets annotated with these dimensions, which are scarce.

\textbf{Relevance to education}: We believe our findings hold promise for educational applications, particularly in personalized QG. Recent work has explored adapting QG to student ability \citep{tomikawa_2024_adpative}. We argue that incorporating narrative control adds another valuable layer to personalization, enabling more targeted and contextually rich QG.

\section{Conclusions}

This work investigates a strategy for controlling both narrative and difficulty attributes in generated QA pairs.
The results offer a preliminary yet promising demonstration of the potential of QG models and the proposed control strategy.
Future efforts could leverage larger datasets with a more balanced distribution of questions across categories to improve the model's control capabilities. 
Additionally, examining the impact of different inference methods on generation would be valuable, especially to address the issue of repetitive outputs observed with beam search. Finally, future research could explore few-shot prompting techniques, providing minimal examples to assess the model's control ability without extensive training.

\section*{Limitations}

While our approach provides promising insights into controllable QG, some limitations should be acknowledged.

First, \textit{the limited representation of question categories across narrative attributes and difficulty levels hinders the model's ability to learn effectively}. FairytaleQA consists of approximately 10k instances. Associating questions with multiple narrative elements and difficulty levels significantly reduces the number of examples per category, limiting the model's ability to learn effectively. For instance, as shown previously in Table~\ref{tab:dist_values_train_val}, \textit{prediction} and \textit{feeling} questions are poorly represented.

Second, \textit{top-k sampling enables control over narrative elements and question difficulty but can lead to undesired hallucinations}. Initially, we experimented with beam search --- a more commonly used technique for QG --- but found it often generated repetitive questions when addressing the same narrative element across varying difficulty levels. Moreover, our findings indicate that the choice of inference method significantly impacts control. For instance, as shown in Section~\ref{sec:results}, the diversity of the generated QA pairs increases at higher difficulty levels. However, this diversity can also produce unintended side effects, such as the hallucinations noted with error analysis.
While hallucinated QA pairs may affect evaluation by inflating perceived difficulty, we believe that reporting such cases was important to reveal potential failure modes of controllable QG systems. Although they may add some noise, these observations help contextualize the results and guide future improvements in model robustness.

Third, the \textit{evaluation relies on simulated learner responses rather than real student data}. While this approach offers scalability and approximations of question difficulty, it may not fully reflect how actual students would respond. Nonetheless, it provides a valuable proxy for assessing the model's behavior, and we believe it still offers meaningful insight into the controllability of QG systems. Future work should explore incorporating real student data to further validate these findings.

\section*{Ethics Statement}

This research involves the automatic generation of QA pairs from narrative texts, incorporating control attributes such as difficulty level and narrative elements. The dataset used, FairytaleQA, consists of human-authored QA pairs from publicly available fairy tales. No personally identifiable or sensitive information is included, ensuring compliance with ethical guidelines for data usage.
The generated QA pairs were evaluated using both automatic metrics and manual inspection to identify potential errors, such as hallucinated content and nonsensical questions. We acknowledge that these models may introduce unintended errors or biases. While this paper does not focus on error mitigation, future work could explore extended human-in-the-loop validation to enhance the reliability of generated QA pairs, particularly in deployment scenarios.

\section*{Acknowledgments}

The authors would like to thank Professor Masaki Uto and Yuto Tomikawa for their helpful clarifications and discussions related to prior work.
This work was financially supported by UID/00027 -- the Artificial Intelligence and
Computer Science Laboratory (LIACC), funded by Funda\c{c}\~{a}o para a Ci\^{e}ncia e a Tecnologia, I.P./ MCTES through national funds.
Bernardo Leite is supported by a PhD studentship (with reference 2021.05432.BD), funded by FCT.

\bibliography{custom}

\appendix

\section{Narrative Control: Semantic Similarity} \label{sec:sem_control}

\begin{table*}[!ht]
\centering
\renewcommand{\arraystretch}{1.2} 
\begin{tabular}{l cccccc c} \hline
\textbf{Data Setup} & \textbf{Char.} & \textbf{Setting} & \textbf{Action} & \textbf{Feeling} & \textbf{Causal} & \textbf{Out.} & \textbf{Pred.} \\ \hline
\textbf{Text $\rightarrow$ QA} & .332& .332& .353& .370& .360& .346& .358\\
\textbf{Nar + Text $\rightarrow$ QA} & .379& .504& .422& .491& .418& .444& .409\\
\textbf{Nar + Dif + Text $\rightarrow$ QA} & .378& .482& .413& .499& .417& .422& .401\\ \hline
\end{tabular}
\caption{\textbf{Narrative Control}: Semantic similarity (BLEURT) between generated and ground-truth questions on the test set by narrative element.
\textbf{Text $\rightarrow$ QA} is used as a baseline to assess whether narrative control helps the generated questions approximate the ground-truth questions.}
\label{tab:results_sem_narrative_control}
\end{table*}

Table~\ref{tab:results_sem_narrative_control} presents the results from the narrative control perspective, measured using BLEURT \citep{sellam_2020_bleurt}.
The goal is to show an improvement in semantic similarity to ground-truth questions when narrative control attributes are incorporated. As observed with ROUGE$_L$-F1 similarity (recall Section~\ref{sec:results}), this trend is observed across all seven narrative labels.
When narrative and difficulty labels are fused, we observe a similar improvement trend, comparable to the incorporation of narrative attributes alone.
These results further support Hypothesis 1 (H1) --- \textit{incorporating narrative attributes will result in generated questions that are more similar to the ground-truth} --- indicating that our method controls the narrative elements underlying the generated questions.

\section{Difficulty-Only vs. Difficulty+Narrative Control} \label{sec:results_en_dif_vs_difnar}

\begin{figure}[!ht]
    \centering
    \includegraphics[width=\columnwidth]{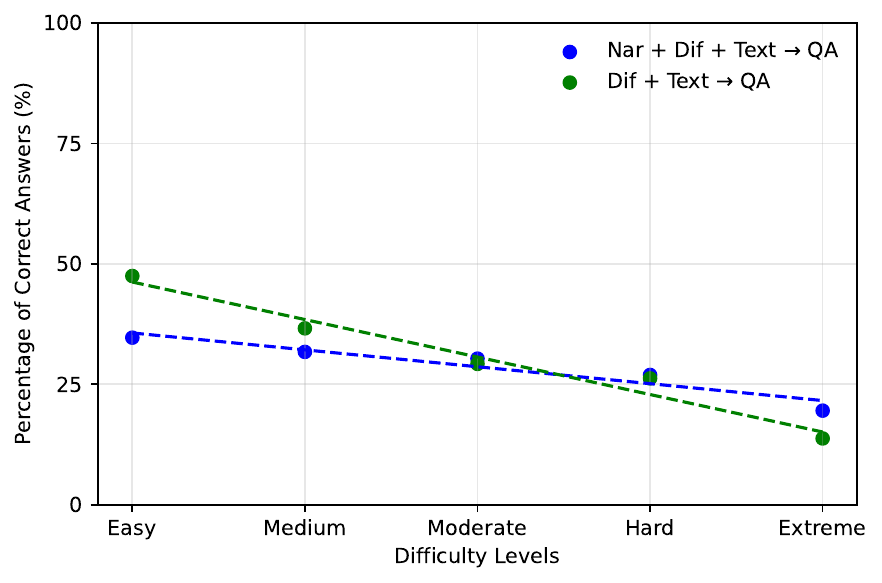}
    \caption{Percentage of Correct Answers by Dif. Level.}
    \label{fig:results_en_dif_vs_difnar}
\end{figure}

To compare difficulty control when operating solely on difficulty versus combining difficulty and narrative attributes, Figure~\ref{fig:results_en_dif_vs_difnar} provides an overview of the performance at each level for both setups. Both setups show the expected trend: the percentage of correct answers decreases as difficulty increases. However, a linear approximation of the observed data points reveals that the decrease is less pronounced when both attributes are combined, though it remains consistent overall.

\section{Additional Linguistic Features Influenced By Control} \label{sec:linguistic_features}
Table~\ref{tab:avg_words} presents the average number of words in the generated question-answer pairs. For generated answers, when only difficulty labels are incorporated, no significant trend is observed. For generated questions, an upward trend is noted, though it is not significant. When narrative and difficulty labels are combined, no trend is observed. Based on these findings, we conclude that the average length of generated question-answer pairs is not influenced by difficulty or narrative control labels in our experiments.
\begin{table}[!ht]
\centering
\begin{tabular}{ccccc}
\hline
\textbf{Data Setup} & & \textbf{Easy} & \textbf{Med.} & \textbf{Extr.} \\ \hline
\multirow{2}{*}{\makecell[c]{\textbf{Dif + Text} \\ $\rightarrow$ \textbf{QA}}} & Q & 10.80 & 11.83 & 12.49 \\
\cline{2-5}
 & A & 7.19 & 8.95 & 8.88 \\ \hline
\multirow{2}{*}{\makecell[c]{\textbf{Nar + Dif + Text} \\ $\rightarrow$ \textbf{QA}}} & Q & 11.81 & 11.62 & 11.70 \\
\cline{2-5}
 & A & 7.42 & 7.96 & 7.61 \\ \hline
\end{tabular}
\caption{Average number of words for generated questions (Q) and answers (A).}
\label{tab:avg_words}
\end{table}

\begin{figure}[!ht]
    \centering
    \includegraphics[width=\columnwidth]{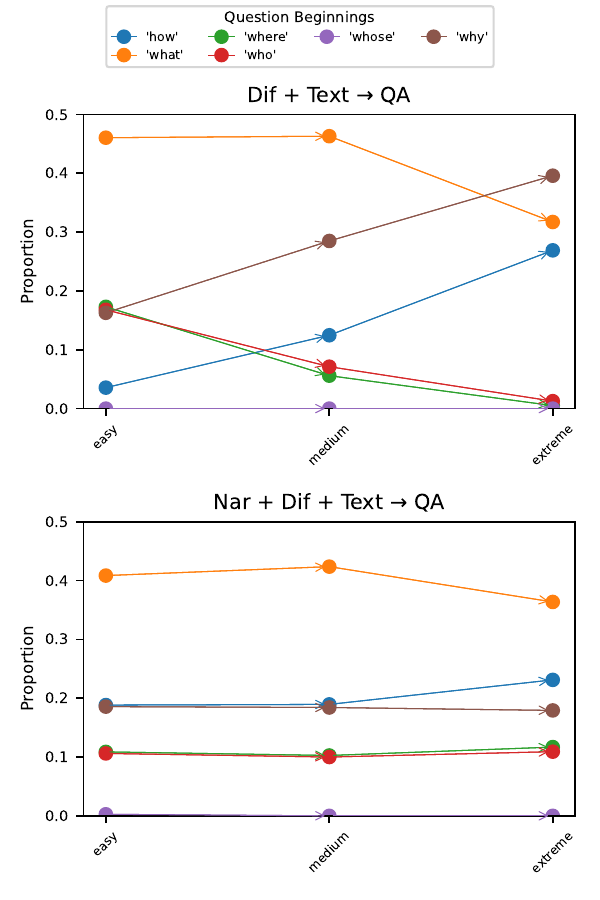} 
    \caption{Proportion of initial interrogative terms in the generated questions (arrowed lines indicate increase/decrease trends).}
    \label{fig:question_begin}
\end{figure}

Figure~\ref{fig:question_begin} illustrates the proportion of initial interrogative terms in the generated questions. When only difficulty labels are used (top chart), higher difficulty levels show an increase in terms like ``why'' and ``how'' and a decrease in terms like ``what'' ``who'' and ``where''. This aligns with expectations, as ``why'' and ``how'' are often linked to questions requiring higher cognitive effort, as described in Bloom's taxonomy \cite{bloom_2002_revised}. When both narrative and difficulty labels are fused (lower chart), the proportion of all interrogative terms is more consistent across difficulty levels. This outcome is expected since this setup aims to control difficulty levels while also demanding for certain narrative elements. In this case, narrative labels are the primarily influence for the choice of interrogative terms (e.g., ``who'' for character-related questions), rather than difficulty labels.

\end{document}